\documentclass[10pt,twocolumn,letterpaper]{article}

\usepackage{cvpr}              

\usepackage[dvipsnames]{xcolor}

\definecolor{cvprblue}{rgb}{0.21,0.49,0.74}
\usepackage[pagebackref,breaklinks,colorlinks,citecolor=cvprblue]{hyperref}

\def\paperID{*****} 
\def\confName{CVPR}
\def\confYear{2024}

\title{The Solution for the CVPR2024 NICE Image Captioning Challenge}

\author{Longfei Huang\textsuperscript{1},
Shupeng Zhong\textsuperscript{1},
Xiangyu Wu\textsuperscript{1,2},
Ruoxuan Li\textsuperscript{1} \\
\textsuperscript{1}{Nanjing University of Science and Technology}
\textsuperscript{2}{Alibaba}\\
}

\begin{document}
\maketitle
\begin{abstract}
This report introduces a solution to the Topic 1 Zero-shot Image Captioning of 2024 NICE : New frontiers for zero-shot Image Captioning Evaluation. In contrast to NICE 2023 datasets, this challenge involves new annotations by humans with significant differences in caption style and content.  Therefore, we enhance image captions effectively through retrieval augmentation and caption grading methods. At the data level, we utilize high-quality captions generated by image caption models as training data to address the gap in text styles. At the model level, we employ OFA (a large-scale visual-language pre-training model based on handcrafted templates) to perform the image captioning task. Subsequently, we propose caption-level strategy for the high-quality caption data generated by the image caption models and integrate them with retrieval augmentation strategy into the template to compel the model to generate higher quality, more matching, and semantically enriched captions based on the retrieval augmentation prompts. Our approach achieves a CIDEr score of 234.11.
\end{abstract}    
\section{Introduction}
\label{sec:intro}
Recently, deep multi-modal learning has acctracted wide \cite{yang2024alignment, yang2023deep, yang2022domfn} and has real applications, with image-text generation being one of the primary applications. Zero-shot image captioning task\cite{LiuLM24, FuSZY24} requires both visual understanding of scenes and the ability to describe scenes in natural language, aiming to generate concise textual descriptions for given images. However, in real-world scenarios, obtaining high-quality human-annotated data is always challenging. Current popular image captioning datasets are mostly collected through web scraping, leading to inconsistent data quality, while caption styles also differ from manually annotated data.

\begin{figure}[t]
	\centering
	\includegraphics[width=\linewidth]{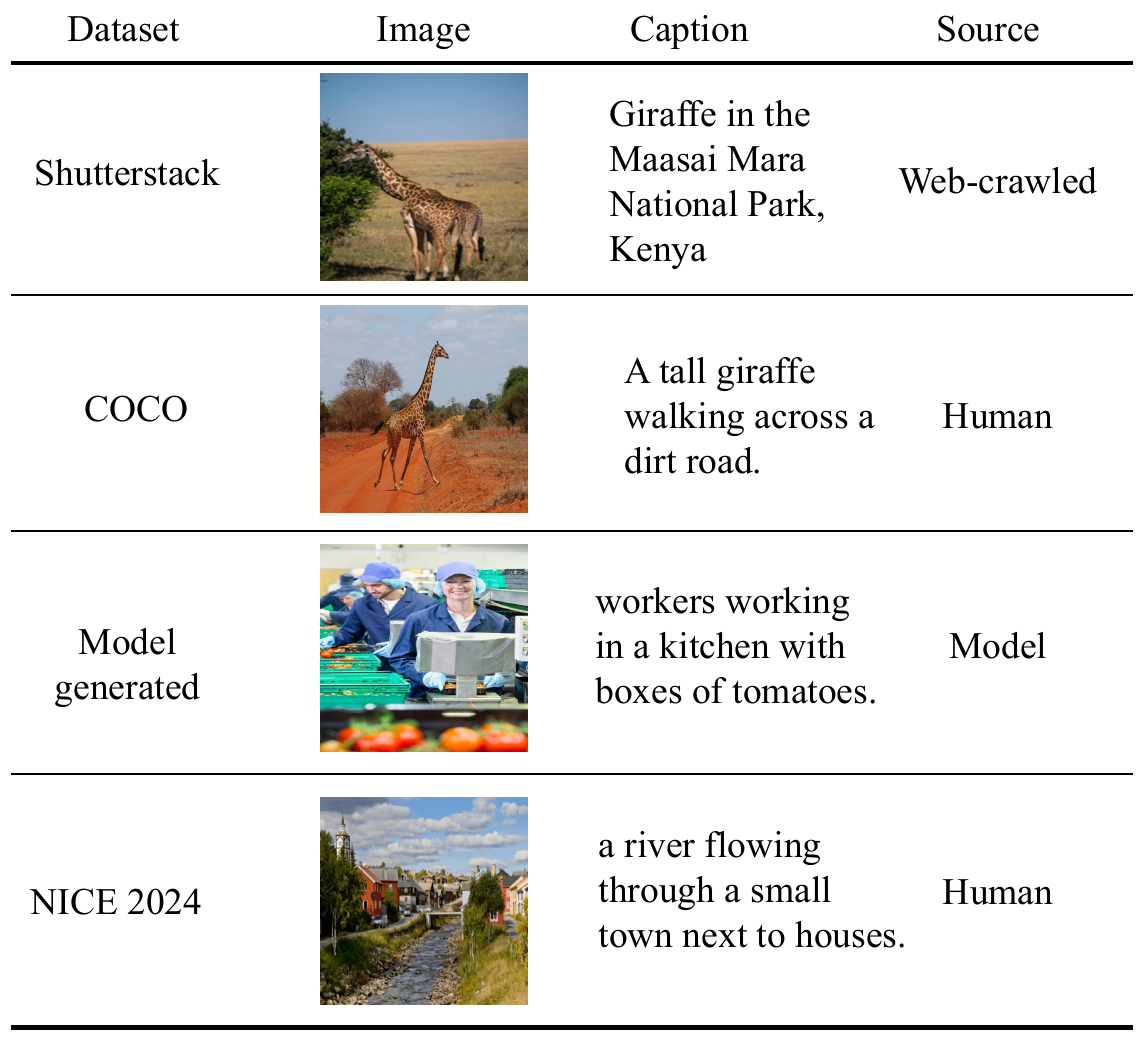}
	\caption{Shutterstock dataset: the web-scraped data, exhibit significant differences in text style compared to manually annotated data. In contrast, the data generated by the model aligns closely with the stylistic characteristics of manually annotated datasets such as COCO and NICE 2024.}\label{fig: dataset}
\end{figure}

The NICE 2024 dataset \cite{NICE} is a zero-shot evaluation dataset composed of images used in the 2023 challenge but with new annotations by humans. It includes approximately 21k images annotated with 5 captions generated by different annotators, providing diversity and accuracy. As shown in Figure \ref{fig: dataset}, the style of captions is similar to manually annotated image caption datasets like COCO \cite{chen2015microsoft}, but differs significantly from web-scraped datasets such as Shutterstock \cite{Shutterstock}. Therefore, learning a consistent representation of vision and language only from large-scale web-scraped data makes it difficult to approach the high-quality results achieved through manual annotation. The COCO dataset is a manually annotated image captioning dataset where each image is provided with five descriptions generated by different humans. However, unlike the COCO dataset, the NICE 2024 dataset contains more complex and diverse visual concepts and types of images.

To address these challenges, we use OFA \cite{WangYMLBLMZZY22} as our base model and leverage externally models generated dataset \cite{model-generated} to effectively expand the model's knowledge. OFA is a large-scale task-agnostic and modality-agnostic universal framework that unifies modalities and tasks using a simple sequence-to-sequence learning framework. Furthermore, we found that data quality is far more important than quantity. Therefore, we use high-quality captions generated by the model as the source of our base dataset.

In addition, we explored some strategies to enhance zero-shot capabilities:(1) We proposed a caption-level strategy that incorporates image-text ranking levels into the template, thus guiding the model to generate captions that are of higher quality and more matching than the prompt. (2) We utilized retrieval-augmentation strategy to integrate knowledge related to input samples into the model, helping the model generate more matching captions by leveraging useful external knowledge. As a result, our approach ranked first on the leaderboard with a CIDEr score of 234.11 and 1st in all other metrics. In the remaining sections of this technical report, we will introduce the detailed architecture of the solution for this challenge.
\section{Related Work}
\label{sec:Related}
\subsection{Vision-language Pre-training Models}

Visual Language Pre-training \cite{0008LSH23,LiXTWYBYCXCZHHZ22} aims to enhance the performance of downstream tasks by pre-training models on a large-scale image-text pair dataset. Visual language pre-training models have shown remarkable success in various multimodal tasks \cite{ChenLYK0G0020,RombachBLEO22}. These works first pre-train a multimodal model on a large-scale image-text dataset and then fine-tune the model for specific tasks. For example, works like BLIP2 \cite{0008LSH23} keep the language model frozen and adjust the visual encoder for the language model. 

There are mainly two types of visual language pre-training models: single-stream models and dual-stream models. Single-stream models refer to models where text and visual features are concatenated together and input into a single transformer block. Dual-stream models refer to models where text and visual features are sent independently to two different transformer blocks. In contrast to the aforementioned architectures, the OFA model formulates pre-training and fine-tuning tasks in a unified sequence-to-sequence abstraction through manual instruction to achieve task-agnostic capabilities.

\subsection{Image Captioning}

Image captioning serves as a crucial link between computer vision (CV) and natural language processing (NLP), aiming to generate contextually and grammatically appropriate textual descriptions for images. With the advent of deep learning technologies, significant progress has been made in this field. Image captioning finds wide-ranging applications, including image and video search, assistive technologies for the visually impaired, as well as automatic content generation for social media and marketing.Typically, image captioning methods utilize pre-trained networks to extract visual features that are then passed to a text decoder to generate the final captions \cite{LuoJSCWHLJ21,TsimpoukelliMCE21}. To bridge the gap between vision and language, other works create a shared latent space for visual and textual information \cite{WangYYDT022,yang2022exploiting}.

The emergence of the CLIP \cite{RadfordKHRGASAM21} marks a turning point in visual language perception. Recent captioning approaches have utilized CLIP to reduce training time and enable zero-shot settings \cite{khandelwal2022simple,tewel2022zerocap}. Zero-shot image captioning aims to generate text descriptions without the need for annotated data, with the goal of explaining images at the expected level of detail while avoiding the inclusion of any erroneous information. This involves the development of algorithms and models that can analyze visual content and generate corresponding textual descriptions. However, zero-shot techniques often lead to lower performance as the generated captions may not align with the desired target style, which is typically determined by the dataset.

\subsection{Vision-Language Retrieval}

Visual language retrieval \cite{wancovlr,YangZXYZY21} is a cross-modal retrieval task that combines computer vision and natural language processing technologies. The primary objective of visual language retrieval is to analyze image content and associated textual descriptions, enabling the retrieval of images from an image library that match a given natural language query or retrieving textual descriptions related to a given image from text data. The goal of visual language retrieval is to create more intuitive and efficient ways for human-machine interaction, such as through image and video search, automatic image captioning, and visual question-answering. \cite{fei2022deecap} proposed a similarity graph reasoning module that relies on graph convolutional neural networks. \cite{ohtechnical} performs k-nearest neighbor (KNN) search using cosine similarity calculated by the CLIP model as the metric. With the development of Transformer-based language understanding, large-scale visual language transformers have sparked deeper modal interactions in retrieval models.


\begin{figure*}[!h]
	\centering
	\includegraphics[width=\linewidth]{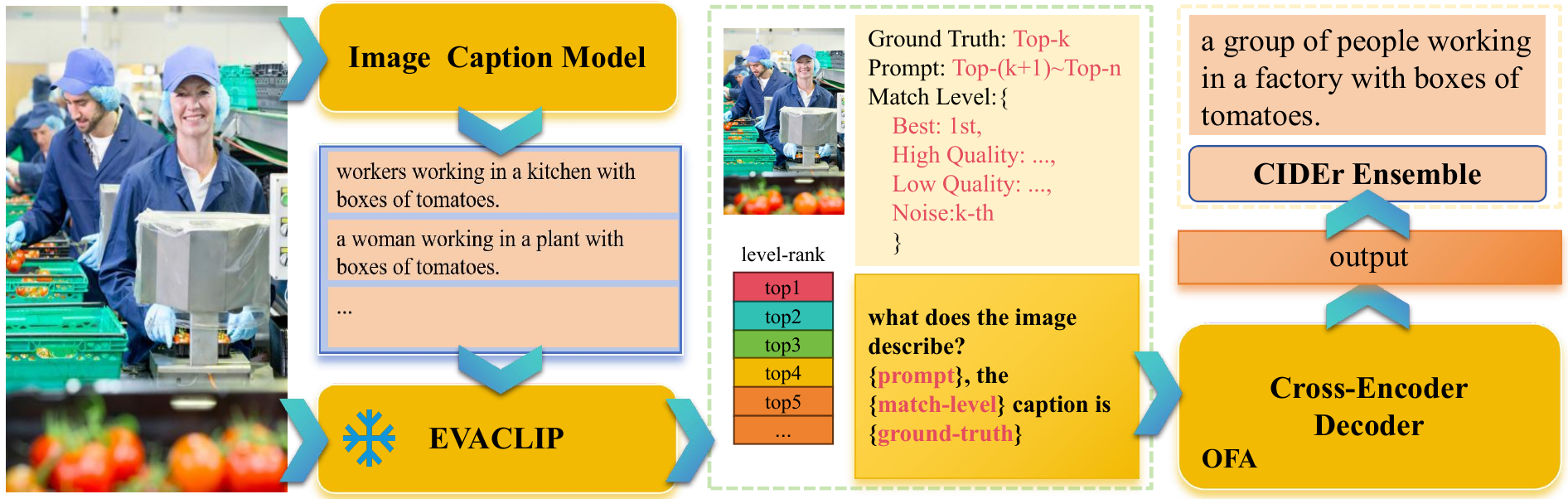}
	\caption{Overall Architecture. Our solution consists of four main stages, which includes Data discovery, Fine-tuning (Retrieval-augmented and Caption-level strategies) and Model-ensemble. The training data are all collected from the models generated dataset.}\label{fig:overview}
\end{figure*}

\section{Methodology}
\subsection{Overall Architecture}

As shown in Figure \ref{fig:overview}, the overall architecture of our solution consists of three parts: Dataset Discovery, Fine-tuning, and Model-ensemble. In the Dataset Discovery phase, multiple captions are generated for test images by the Image Caption Model, which are then retrieved and ranked to obtain high-quality data, as detailed in Secs.\ref{sec:data discovery}. The Fine-tuning phase primarily enhances the model's zero-shot performance through retrieval augmentation and caption-level strategies, enabling the model to learn more comprehensive knowledge, as elaborated in Secs.\ref{sec:Retrieval-augmented} and Secs.\ref{sec:caption-level}. Model-ensemble is the final stage, where the CIDEr-Ensemble trick is employed to improve the model's robustness in zero-shot image captioning tasks. We will provide a detailed explanation in Sec.\ref{sec:Model-ensemble}.

\begin{table}[!ht]
\centering
\begin{tabular}{cccc}
\toprule[1.5pt]
\# & \textbf{Method}  &  \textbf{Num data}  &  \textbf{CIDEr} \\ 
\midrule 
1 & web-crawled fine-tuned & 115k & 155+ \\
2 & COCO fine-tuned & 0 & 171+ \\
3 & Model-gen fine-tuned & 115k & 188+ \\
\bottomrule[1.5pt]
\end{tabular}
\caption{Comparing the fine-tuning results of various types of datasets, the quality of data generated by the model is better when compared to the COCO dataset and web-crawled data. `` COCO fine-tuned '' uses the officially released model weights. `` Web-crawled '' primarily refers to the data retrieved from Laion-5B and Shutterstock.`` Model-gen '' denotes the data that has been reannotated using image caption models on the web-crawled data.`` Web-crawled ''and `` Model-gen ''use pre-training model weights} \label{tab:table-1}
\end{table}

\subsection{Data discovery} \label{sec:data discovery}

As shown in Table \ref{tab:table-1}, the OFA pre-trained model performs poorly on the NICE 2024 dataset even after fine-tuning on web-crawled data, and its performance remains unsatisfactory even after fine-tuning on manually annotated COCO dataset. The web-crawled data mainly comes from Shutterstock \cite{Shutterstock} and Laion-5B \cite{schuhmann2022laion}, which are images retrieved by the CLIP model and KNN algorithm \cite{ohtechnical, IscenFS23} that are similar to those in the NICE 2024 dataset. This indicates that simply fine-tuning with a large amount of similar image-text pair data does not meet the evaluation requirements of the NICE dataset; the quality of the caption data is far more important than the quantity.

Furthermore, we observed significant differences in text description styles between the web-crawled data and the manually annotated data, with the latter containing a narrower range of knowledge due to its scarcity. To address this issue, we attempted to reannotate the web-crawled data using different image caption models (including OFA \cite{WangYMLBLMZZY22}, BLIP2 \cite{0008LSH23}, and mPLUG \cite{LiXTWYBYCXCZHHZ22}). To maintain the style of text descriptions, all models were fine-tuned with COCO weights. After fine-tuning on the reannotated data, the model's performance significantly improved, validating the effectiveness of training with model-generated captions.

Therefore, we decided to use the images from the NICE dataset along with captions generated by official models \cite{model-generated} as our dataset. To ensure high-quality training data while reducing training costs, instead of selecting some image-text pairs similar to NICE images and mixing them with NICE data for training, we directly utilized the clean NICE dataset. This approach also leads to the generation of higher quality and better matching captions by the model.

\begin{figure}[t]
	\centering
	\includegraphics[width=\linewidth]{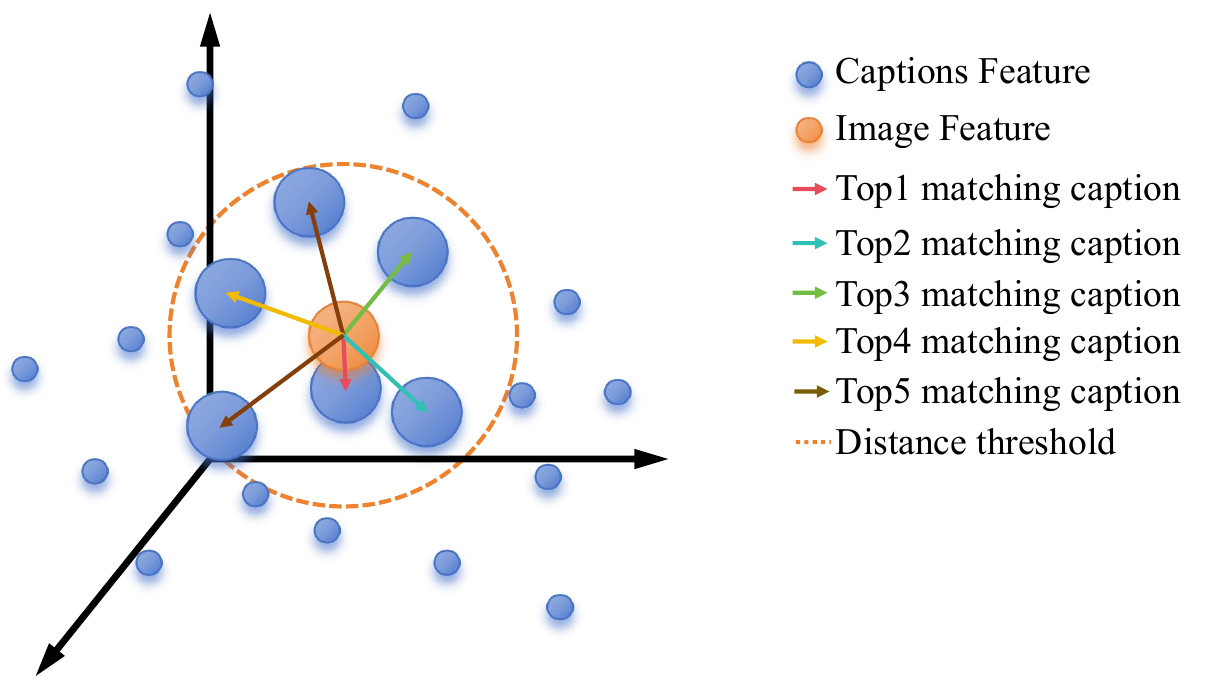}
	\caption{Establishing a dataset using visual language retrieval with the EVA-CLIP model and Adaption Re-ranking method. }\label{fig:data-discover}
\end{figure}

Additionally, to further ensure the quality of captions in the dataset, we introduced a retrieval module based on the EVA-CLIP \cite{sun2023eva} model combined with the Adaption Re-ranking method \cite{yang2023self} to filter the data, selecting the \textit{top-n} image-caption pairs with high image-caption matching as the foundational dataset. As shown in Figure \ref{fig:data-discover}. Subsequently, we designate the \textit{top-k} data for each image from the foundational dataset as ground-truth, proportionally dividing it into training and validation sets, while the data from \textit{(k+1)-th} to \textit{n-th} is used to construct a mini-knowledge-base.

Given the encoded image query embedding \(M_q=\left\{w_0^I\right\}\) and M candidate encoded textual embeddings \(M_c={\left\{w_i^T\right\}}_{i=1}^M\) processed in parallel. Firstly, compute the query-candidate similarity as follows:
\[S=\left(M_c\right){\left(M_q\right)}^{'}\in \mathbb{R}^{M\times1} \tag{1}\]
To simplify computations, broadcast the query-candidate similarity metric \(S\) into \(\tilde{S}={\left\{S\right\}}_{i=1}^D \in \mathbb{R}^{M\times D}\). Then perform SVD on \(M_c\) under the constraint of similarity \(S\) , yielding:
\[M_s=M_c \odot \tilde{S}=U \Sigma V^T \tag{2}\]
By utilizing the column vector \(U_1 \in \mathbb{R}^{M\times1}\) of \(U\) \textit{w.r.t.} the largest singular value of \(M_s\), we can get the principal vector of the encoded candidate captions \(M_q^{*} \in \mathbb{R}^{1\times D}\) , which is used as the re-ranking query:
\[M_q^{*}=U_1^{'}M_c \tag{3}\]
Finally, incorporate the k-reciprocal encoding algorithm \cite{QinGBQG11} to get the refine re-ranking results:
\[S^{*}=k{-}reciprocal\left(M_q^{*}, M_c\right) \tag{4}\]
Based on the new similarity matrix \(S^{*}\), we can get the \textit{top-n} captions that match the image query.

\subsection{Retrieval-augmented} \label{sec:Retrieval-augmented}

The retrieval augmentation strategy provides a mini knowledge base for each image-text pair during training. By explicitly aligning the information of images with the knowledge in the knowledge base, the model helps learn visual features such as objects, attributes, and relationships of the images. We found that quality is more important than quantity of data; therefore, we use the EVA-CLIP model to query the matching level between images and captions and then employ the CIDEr-Ensemble trick to select the \textit{top-k} caption data for each image as ground truth. The remaining caption data is concatenated to form a mini knowledge base, resulting in a diverse and high-quality caption dataset with comprehensive information. 

For each ground truth of each image, \textit{$n$} relevant retrieval knowledge entries are randomly selected from the knowledge base as prompts and inserted into the template of the OFA model to obtain the template: ``\textit{What does the image describe?} \{prompt\}\textit{, the caption is}''. By utilizing the retrieved knowledge, the model can generate accurate captions. However, since all textual data is generated by the model, it may lead to error accumulation and quality bottlenecks in caption generation. We will discuss our solution in detail in the following Sec.\ref{sec:caption-level}.

\begin{figure}[t]
	\centering
	\includegraphics[width=\linewidth]{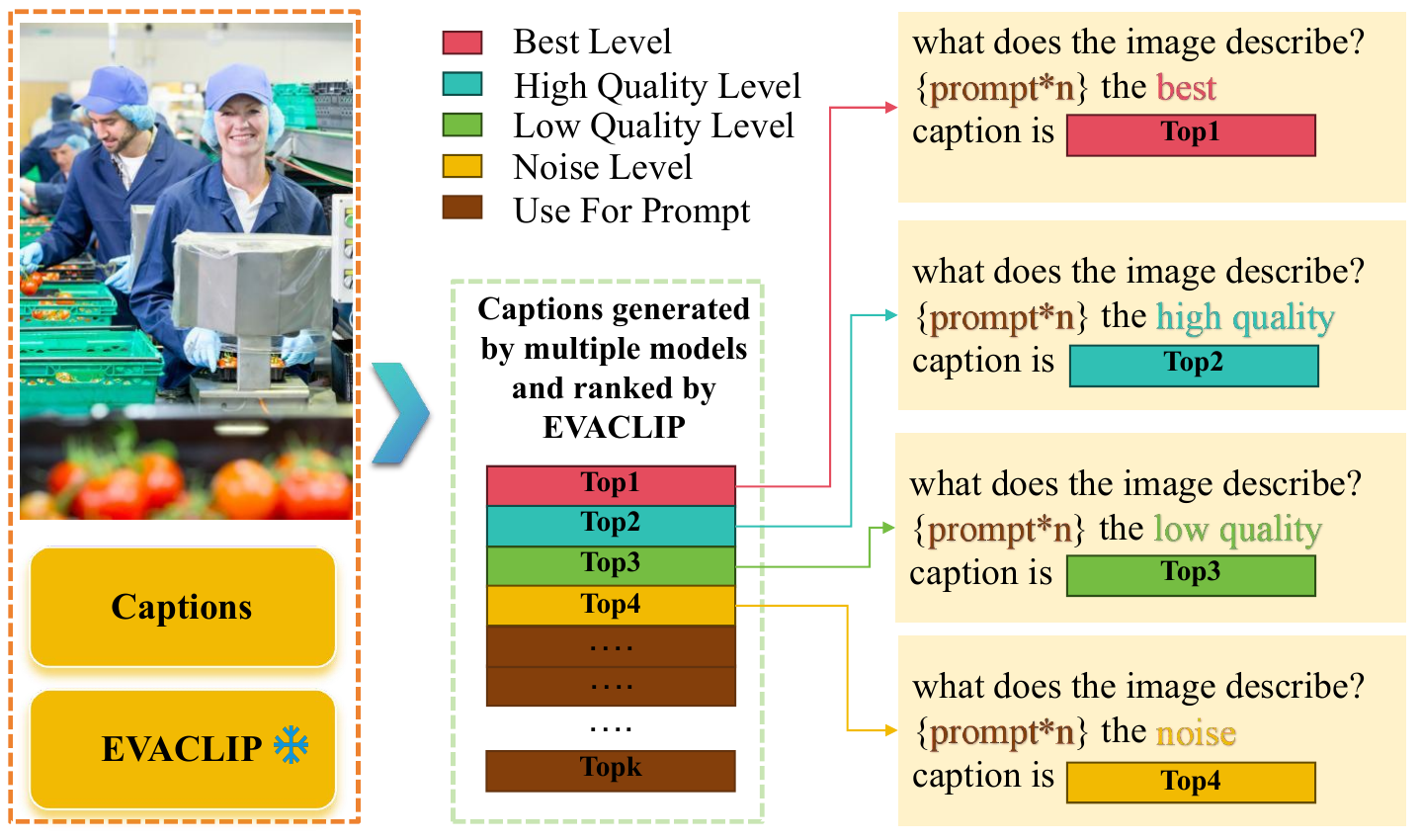}
	\caption{Caption-level is utilized in fine-tuning stages. }\label{fig:caption-level}
\end{figure}

\subsection{Caption-level} \label{sec:caption-level}

Caption-level strategy provides different levels of caption quality hints for visual language models during the fine-tuning stage. As shown in Figure \ref{fig:caption-level}. In the fine-tuning stage, we define \textit{$n$} caption levels by matching the selected \textit{top-k} caption data from high to low quality, and by incorporating caption level hints into the model's template, the model can independently learn representations of captions of different qualities. During inference, the caption level is fixed at the highest level, forcing the model to generate high-quality and highly matched subtitles.

Different from similarity-bucket strategy \cite{KangMLR23, wu2023solution}, our caption-level method does not calculate the similarity for all image-text pairs and then sort them to divide into different similarity thresholds for each bucket. Caption-level involves ranking different captions for each image. similarity-bucket strategy is more suitable for data obtained through web crawling, as the quality of such data varies widely. By assigning different similarity thresholds to all image-text pairs, similarity-bucket strategy can effectively distinguish between low-quality and high-quality image-text pairs. However, our data comes from model generation and has been filtered through sorting, with each image having multiple captions, all of which are of relatively higher quality corresponding to the image. It is difficult for similarity-bucket strategy to effectively divide the data. The caption-level strategy can effectively classify different captions of the same image, enabling the model to better learn higher quality and more matched captions. During the fine-tuning process and the inference stage, we insert the caption-level into the template of the OFA model, resulting in the template: ``\textit{What does the image describe? the} \{level\} \textit{caption is}''.

Additionally, the prompts in the constructed mini-knowledge base are generally of lower quality compared to the selected ground truth. By employing a caption-level strategy to classify levels, we can encourage the model to learn to generate captions that are of higher quality, better matching, and more richly accurate than the prompts. During the inference stage, when we insert the highest level caption as the prompt into the template, the model may generate new captions of higher quality, surpassing the currently filtered highest level caption. This effectively addresses the issues of error accumulation in model-generated captions and the bottleneck in the quality of generated caption content.

\subsection{Model-ensemble} \label{sec:Model-ensemble}

In the final Model-ensemble stage, we employ the CIDEr-ensemble trick to integrate multiple sets of results generated by the models. Specifically, we fine-tune different models with varying weights or infer \textit{$n$} prediction results obtained from different prompt combinations. Each prediction is evaluated for CIDEr score with the other \textit{$n-1$} captions, and the caption corresponding to the highest score is selected as the final prediction result. In the fine-tuning stage, the model weights we use include OFA-Large and OFA-Huge weights fine-tuned on the COCO dataset, as well as OFA-Large weights pre-trained through contrastive learning on web-crawled data \cite{wu2023solution}. The prompt combinations used for inference consist of a varying number of textual description sentences selected from the constructed dataset.

\section{Experiments}
\subsection{Implementation Detail}

\textbf{Dataset}. Following the method outlined in Sec.\ref{sec:data discovery}, we filtered out the \textit{top10} image-caption pairs with the highest image-caption matching scores from the 60 captions generated by the official model for each image, resulting in a foundational dataset of 200k data points. Among these, the \textit{top4} image-caption pairs were further divided into training and validation sets at a ratio of 10:1, while the data pairs ranked 5th to 10th were used as the mini-knowledge base. Subsequently, through CIDEr-ensemble on the foundational dataset, we obtained the optimal \textit{top3} image-text pairs as the final prompts for the model's inference process.

\textbf{Model}. We utilized OFA as the foundational image caption model and set the manual template of OFA to: ``\textit{What does the image describe? {prompt}, the {level} caption is}''. For each image, the number of prompts in the corresponding knowledge base was set to 1, 2, and 3. During the fine-tuning stage, we defined four levels of captions: ``\textit{ best match }'', ``\textit{ high quality }'', ``\textit{ low quality }'', and ``\textit{ noise }'', aligned with the \textit{1st-4th} results of the \textit{top4} image caption retrieval. Concurrently, we employed various OFA model pre-training weights, including those of the OFA-Huge and OFA-Large models fine-tuned on the COCO dataset, as well as the OFA-Large model weights pre-trained on web-crawled data using contrastive learning. This approach enabled us to acquire diverse models for executing model ensemble trick. In subsequent experiments, unless specified otherwise, the default loading of weights was from the OFA-Large model fine-tuned on the COCO dataset, with a quantity of 3 descriptive sentences in the prompt.

\textbf{Fine-tuning setup}. We employed \textit{NVIDIA RTX 3090$\times$4} with an initial learning rate of \textit{1e-5}, input image size of \textit{480$\times$480}, batch size of \textit{16}, input text length of \textit{20}, maximum training epochs set to \textit{5}, and saved the model weights with the lowest validation set loss as the fine-tuning result. All other hyperparameters followed the default settings of the OFA model.

\textbf{Evaluation and Model-ensemble setup}. During inference, prompts continue to utilize the same number of relevant sentences as in the fine-tuning stage, but this knowledge is drawn from the top-performing data in the foundational dataset (defaulting to the \textit{top3}). Models with CIDEr scores surpassing 220 points have their predictions integrated into the final result using the CIDEr-ensemble trick.

\begin{figure}[t]
	\centering
	\includegraphics[width=\linewidth]{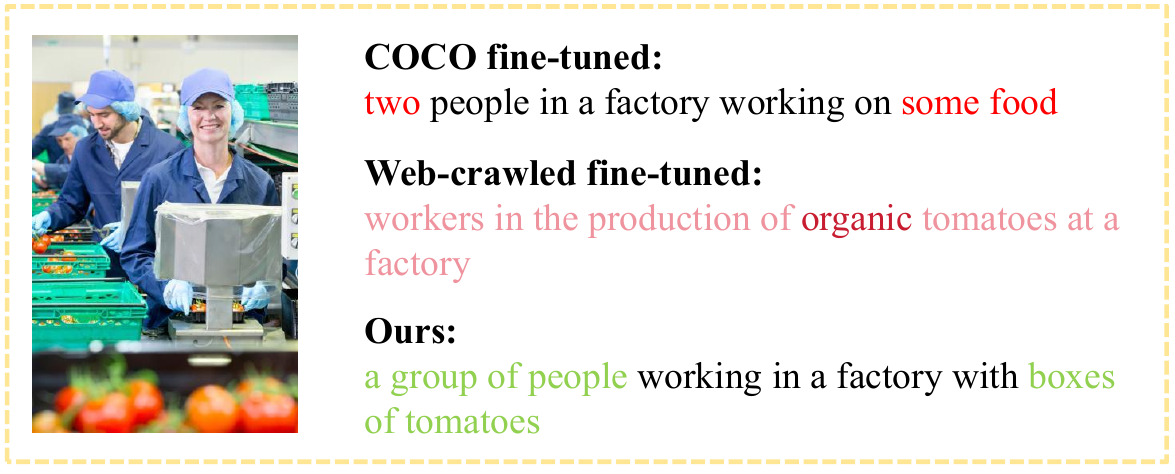}
	\caption{A comparison of prediction results between fine-tuning directly on the COCO dataset, data crawled from the web, and our method. COCO dataset has a relatively limited knowledge scope, and ``COCO fine-tuned'' often results in vague descriptions (such as ``some food'') that fail to accurately predict scenes and object categories. While web-crawled data encompasses a broader range of knowledge, ``web-crawled fine-tuned'' can accurately predict object categories but may introduce fabricated information, leading to misleading terms (such as ``organic tomatoes''), thereby diminishing the practicality of the model. In contrast, our approach not only generates detailed descriptions but also avoids misleading terms, significantly enhancing the model's performance and practicality.} \label{fig:result}
\end{figure}

\subsection{Main Result}

As shown in Figure \ref{fig:result}, compared to fine-tuning directly on the COCO dataset or data crawled from the web, our method yields more accurate and matching predictions that closely resemble manual annotations.

\textbf{Effectiveness of data discovery}. As shown in Table \ref{tab:table-1}, the results of fine-tuning pre-trained OFA on various types of datasets and evaluating on the NICE dataset demonstrate that training the model to generate captions helps retain knowledge while effectively maintaining the high-quality human-annotated textual style. Furthermore, we explored the impact of data quantity and quality on the results. As depicted in Table \ref{tab:table-2}, the fine-tuning performance with more data (\textit{top10} data) showed a decline compared to the fine-tuning results with higher-quality data (\textit{top4} data). Therefore, the quality of data is more critical than the quantity of data, leading us to select higher-quality data.

\begin{table}[!ht]
\centering
\begin{tabular}{cccc}
\toprule[1.5pt]
\# & \textbf{Method}  &  \textbf{Num data}  &  \textbf{CIDEr} \\ 
\midrule 
1 & top10 data fine-tuned & 200k & 206+ \\
2 & top4 data fine-tuned & 80k & 208+ \\
\bottomrule[1.5pt]
\end{tabular}
\caption{Comparing the impact of data quantity and quality on model performance.} \label{tab:table-2}
\end{table}

\begin{table}[!ht]
\centering
\begin{tabular}{cccc}
\toprule[1.5pt]
\# & \textbf{Method}  &  \textbf{strategy}  &  \textbf{CIDEr} \\ 
\midrule 
1 & top10 data fine-tuned & Similarity-bucket & 207+ \\
2 & top10 data fine-tuned & Caption-level & 206+ \\
3 & top4 data fine-tuned & Similarity-bucket & 206+ \\
4 & top4 data fine-tuned & Caption-level & 208+ \\
\bottomrule[1.5pt]
\end{tabular}
\caption{Comparison between similarity-bucket strategy and Caption-level strategy.} \label{tab:table-3}
\end{table}

\textbf{Effectiveness of Caption-level}. As shown in Table \ref{tab:table-3}, as the data quality improves, the effectiveness of similarity-bucket strategy for training with noisy data decreases. similarity-bucket strategy, which grades all data together, is more suitable for web-crawled data with significant differences in data quality. On the other hand, the caption-level strategy, which grades different captions for the same image, can effectively adapt to the cleanliness of high-quality data, thus assisting the model in generating higher-quality prediction results.

\textbf{Effectiveness of Retrieval-augmented}. We found that the Retrieval-augmented strategy can effectively enhance model performance, and the more knowledge contained in the prompt, the more beneficial it is for model predictions. As shown in Table \ref{tab:table-4}, the more knowledge included in the prompt, the better the model generation results. However, once a sufficient amount of knowledge is included, the performance improvement of the model starts to plateau.

\begin{table}[!ht]
\centering
\begin{tabular}{cccc}
\toprule[1.5pt]
\# & \textbf{Method}  &  \textbf{Num text}  &  \textbf{CIDEr} \\ 
\midrule 
1 & top4 data fine-tuned & 1 & 227.16 \\
2 & top4 data fine-tuned & 2 & 230.18 \\
3 & top4 data fine-tuned & 3 & 230.98 \\
4 & Model-ensemble & 0 & 234.11 \\
\bottomrule[1.5pt]

\end{tabular}

\caption{Comparing the impact of the number of relevant description sentences in the prompt on model performance.} \label{tab:table-4}
\end{table}

\textbf{Effectiveness of Model-ensemble}. We used the CIDEr-ensemble trick to optimize captions with CIDEr scores greater than 220 from models fine-tuned under different strategies, ultimately increasing the CIDEr score from over 230+ to 234+.

\section{Conclusion} 

This report summarizes our solution to the zero-shot image captioning challenge, which includes two core strategies: Retrieval-augmented strategy and Caption-level strategy. Our solution demonstrates that the Caption-level strategy can effectively enhance the generation performance of models on high-quality data. Furthermore, both the Retrieval-augmented strategy and Caption-level strategy can effectively mitigate potential issues such as error accumulation from using model-generated captions. The final competition results validate the effectiveness of our solution.

Moreover, our experimental results demonstrate that utilizing Retrieval-augmented strategy and Caption-level strategy can improve model performance through iterative model updates even without the addition of new real data, by leveraging existing knowledge across models. Interestingly, our experiments show that providing captions of the current highest rank (highest quality, highest match) as prompts enables the model to surpass this bottleneck and generate captions of higher quality and better match. The question arises: can repeating this method enable pre-training models to achieve self-iteration and continuous evolution? This will be a focus of our future exploration.
{
    \small
    \bibliographystyle{ieeenat_fullname}
    \bibliography{main}
}


\end{document}